\documentclass[10pt,twocolumn,letterpaper]{article}

\usepackage{cvpr}
\usepackage{times}
\usepackage{epsfig}
\usepackage{graphicx}
\usepackage{amsmath}
\usepackage{amssymb}
\usepackage[normalem]{ulem}
\usepackage{afterpage}
\usepackage{booktabs}
\usepackage{color}
\usepackage{verbatim}
\usepackage{soul}
\usepackage{xspace}
\usepackage{comment}
\usepackage{array}
\usepackage{multirow}
\usepackage{url}
\usepackage{epigraph}
\usepackage[toc,page]{appendix}
\usepackage{pifont}

\newcommand{\xmark}{\ding{55}}

\usepackage[pagebackref=true,breaklinks=true,letterpaper=true,colorlinks,bookmarks=false]{hyperref}

 \cvprfinalcopy 
\def\cvprPaperID{1426} 

\ifcvprfinal\fi
\begin{document}

\title{Shapes and Context: In-the-Wild Image Synthesis \& Manipulation}
\author{Aayush Bansal \quad\quad Yaser Sheikh \quad\quad Deva Ramanan\\
Carnegie Mellon University\\
{\tt\small{\{aayushb,yaser,deva\}@cs.cmu.edu}}
}

\maketitle

\begin{abstract}
\vspace{-0.4cm}
We introduce a data-driven approach for interactively synthesizing in-the-wild images from semantic label maps. Our approach is dramatically different from recent work in this space, in that we make use of no learning. Instead, our approach uses simple but classic tools for matching scene context, shapes, and parts to a stored library of exemplars. Though simple, this approach has several notable advantages over recent work: (1) because nothing is learned, it is not limited to specific training data distributions (such as cityscapes, facades, or faces); (2) it can synthesize arbitrarily high-resolution images, limited only by the resolution of the exemplar library; (3) by appropriately composing shapes and parts, it can generate an exponentially large set of viable candidate output images (that can say, be interactively searched by a user). We present results on the diverse COCO dataset, significantly outperforming learning-based approaches on standard image synthesis metrics. Finally, we explore user-interaction and user-controllability, demonstrating that our system can be used as a platform for user-driven content creation.

\end{abstract}


\section{Introduction}
\label{sec:intro}

\begin{figure*}[t]
  \centering
  \includegraphics[width=\linewidth]{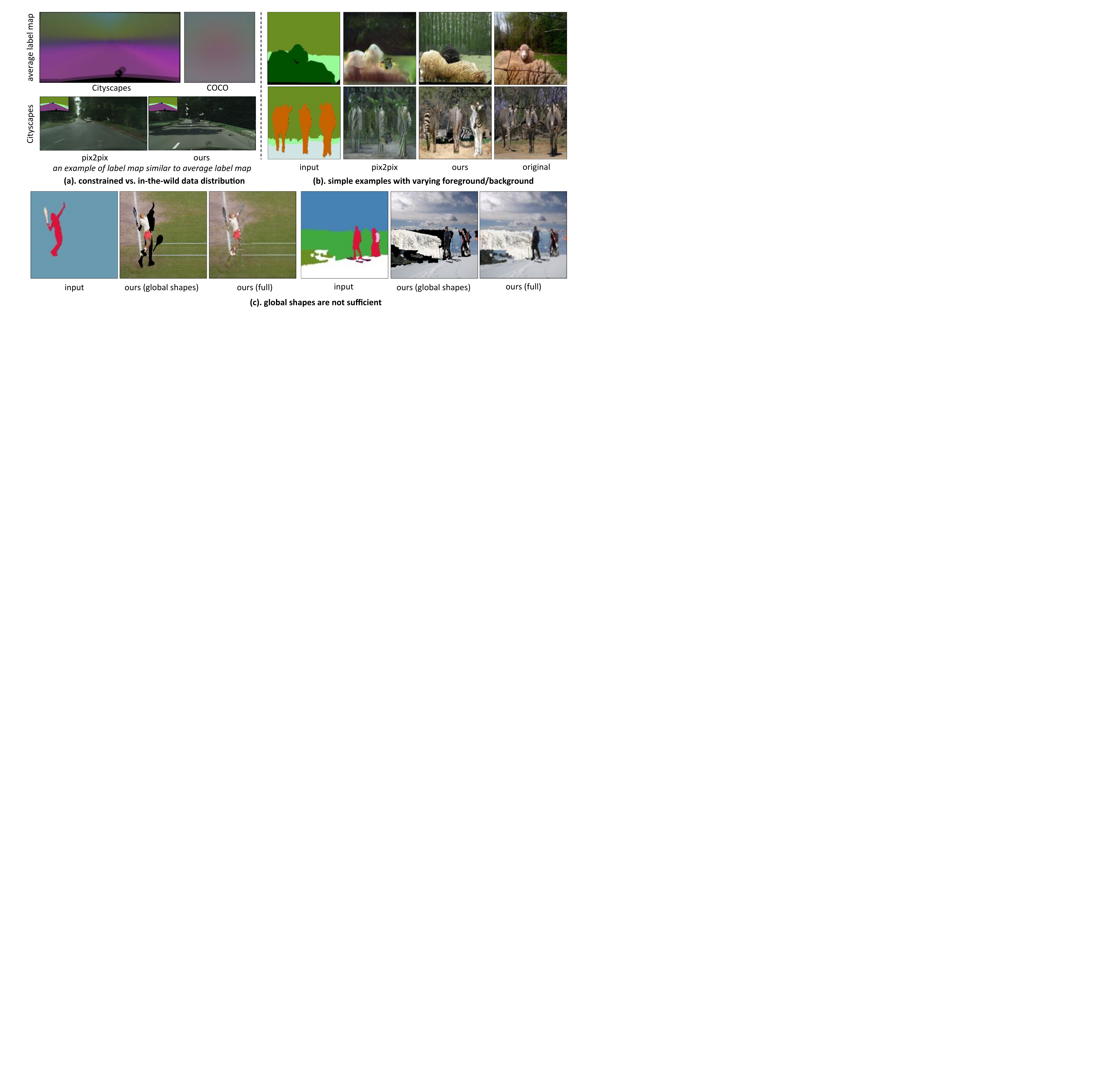}
  \caption{\textbf{Limitations of current approaches for images synthesis: } \textbf{(a)} Current image synthesis models tend to be trained on datasets with somewhat limited diversity, such as Cityscapes~\cite{Cordts2016Cityscapes}, faces~\cite{Smith2013ExemplarBasedFP}, or facades~\cite{Tylecek13}. For example, the average label mask for Cityscapes~\cite{Cordts2016Cityscapes} clearly reveals redundant structure such as a car hood, road, and foliage. In contrast, the average image for COCO~\cite{MSCOCO-2014} is much less structured, suggesting it is a more varied dataset. \textbf{(b)} Indeed, we train state-of-the-art neural architectures~\cite{pix2pix2016,wang2017high} on COCO and observe poor convergence (even after a month of training!) resulting in a mode collapsed and averaged outputs. \textbf{(c)} In contrast, our simple matching-based approach is able to synthesize realistic image content by matching to exemplar shapes. In order to generate high-quality images, we find it crucial to encode scene context and part deformation in the matching process - i.e., matching global shapes alone will produce poor images with missing regions due to shape mismatches. }
  \label{fig:concerns}
\end{figure*}

We introduce a data-driven approach for interactively synthesizing diverse images from semantic label maps. Specifically, we seek to design a system for \emph{in-the-wild} image synthesis that is controllable and interpretable. While content creation is a compelling task in of itself (a classic goal of computer graphics), image synthesis is also useful for generating data that can be used to train discriminative visual recognition systems~\cite{hoffman2018cycada}. Synthesized data can be used to explore scenarios that are difficult or too dangerous to sample directly (e.g., training an autonomous perception system on unsafe urban scenes~\cite{huang2017expecting}). Figure~\ref{fig:teaser_fig} shows images synthesized using our approach, where the input is a semantic label map.

{\bf Parametric vs Nonparametric:} Current approaches for image synthesis and editing can be broadly classified into two categories. The first category uses parametric machine learning models. The current state-of-the-art~\cite{Recycle-GAN,chen2017photographic,pix2pix2016,wang2017high} relies on deep neural networks~\cite{lecun2015deep} trained with adversarial losses (GANs)~\cite{GoodfellowPMXWOCB14} or perceptual losses~\cite{johnson2016perceptual} to create images. These approaches work remarkably well when trained on datasets with somewhat limited diversity, such as Cityscapes~\cite{Cordts2016Cityscapes}, faces~\cite{Smith2013ExemplarBasedFP}, or facades~\cite{Tylecek13}. It is unclear how to extend such approaches for ``in-the-wild'' image synthesis or editing: parametric models trained on one data distribution (e.g. Cityscapes) do not seem to generalize to others (e.g. facades), a problem widely known as dataset bias~\cite{CVPR11_Torralba}. The second category of work~\cite{Efros03,Efros:2001,Hertzmann:2001,lalonde-siggraph-07,russell2009segmenting} uses non-parametric nearest neighbors to create content. These approaches have been demonstrated on interactive image editing tasks such as object insertion~\cite{lalonde-siggraph-07} or scene completion~\cite{Hays:2007}. Though a large inspiration~\cite{Hertzmann:2001} for our own work, such approaches have interestingly fallen out of favor in recent history. 

{\bf Does more data help?} A peculiar property of many parametric synthesis methods is that they do better with {\em less} data~\cite{Recycle-GAN,GoodfellowPMXWOCB14,pix2pix2016,RadfordMC15,han2017stackgan,CycleGAN2017}. The culprit seems to be that such methods don't do well on diverse training sets, and in practice larger training sets tend to be diverse. This is in contrast with truly non-parametric methods that do better with {\em more} data~\cite{Hays:2007}. Figure~\ref{fig:concerns}-(a)-(b) highlights the differences between limited and diverse datasets, using illustrative examples of Cityscapes~\cite{Cordts2016Cityscapes} and COCO~\cite{MSCOCO-2014}. While parametric methods do well on limited data distributions, they struggle to perform on diverse datasets. Recent works~\cite{brock2018large,park2019SPADE} have attempted to overcome this challenge by using enormously large model sizes and massive compute (see concurrent work from Park et al~\cite{park2019SPADE} for more discussion on parametric approaches).

{\bf Composition by parts:} In this work, we make three observations that influence our final approach;  (1) humans can imagine {\em multiple} plausible output images given a particular input label mask. We see this rich space of potential outputs as a vital part of the human capacity to imagine and generate. Most parametric networks tend to formulate synthesis as a one-to-one mapping problem, and so struggle to provide diverse outputs (a phenomena also known as mode collapse). Important exceptions include~\cite{pixelnn,chen2017photographic,Ghosh2017MultiAgentDG,zhu2017toward} that generated multiple outputs by employing various modifications.  (2) Visual scenes are exponentially complex due to many possible compositions of constituent objects and parts. It is tempting to combine both observations, and generate multiple outputs by composing scene elements together. But these compositions cannot be arbitrary - one cannot freely swap out a face with a wheel, or place a elephant on a baseball field.  To ensure consistency, our matching process makes use of implicit {\em contextual} semantics present in the library of exemplar label masks.  (3) Given an exemplar set of sufficient size, nearest neighbor methods may still perform well with simple features that are not learned (e.g., pixel values). We combine our observations to construct an image synthesis system that relies on exemplar shapes and parts using simple pixel features.

\begin{figure}
  \centering
  \includegraphics[width=\linewidth]{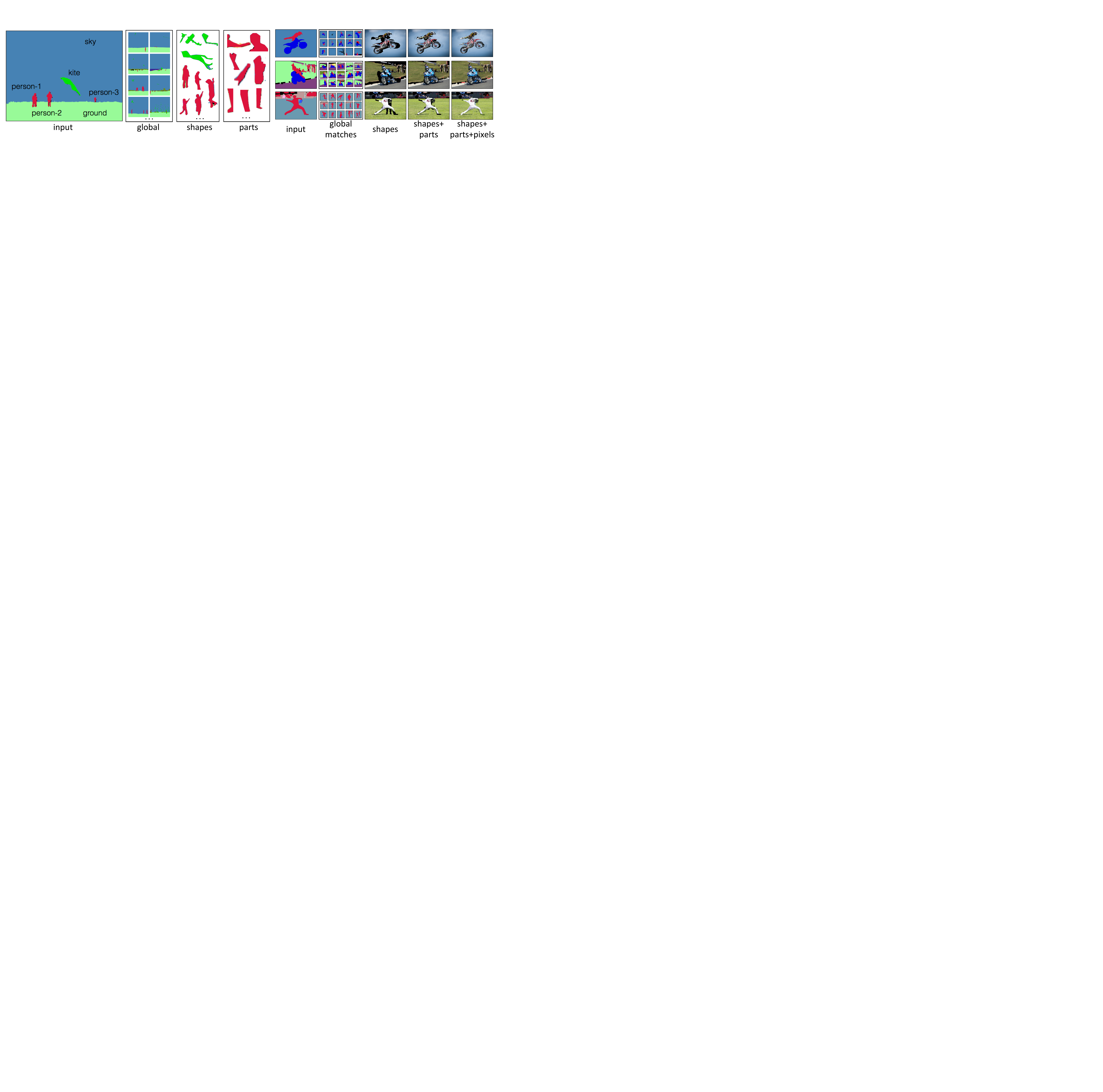}
  \caption{\textbf{Overview: }  Given a label map (left), our goal is to generate multiple plausible images. We make use of hierarchical matching for efficient retrieval of images, shapes, and parts}
  \label{fig:overview2}
\end{figure}

\noindent\textbf{Our Contributions:} (1) We study the problem of visual content creation and manipulation for in-the-wild settings, and observe that reliance on parametric models lead to averaged or mode-collapsed outputs; (2) we present an approach that utilize shapes and context to generate images consisting of rigid and non-rigid objects in varied backgrounds, and different environmental and illumination conditions; (3) we demonstrate the controllable and interpretable aspects of our approach that enables a user to influence the generation and select examples from many outputs.


\section{Background}
\label{sec:background}

\begin{figure*}[t]
  \centering
  \includegraphics[width=\linewidth]{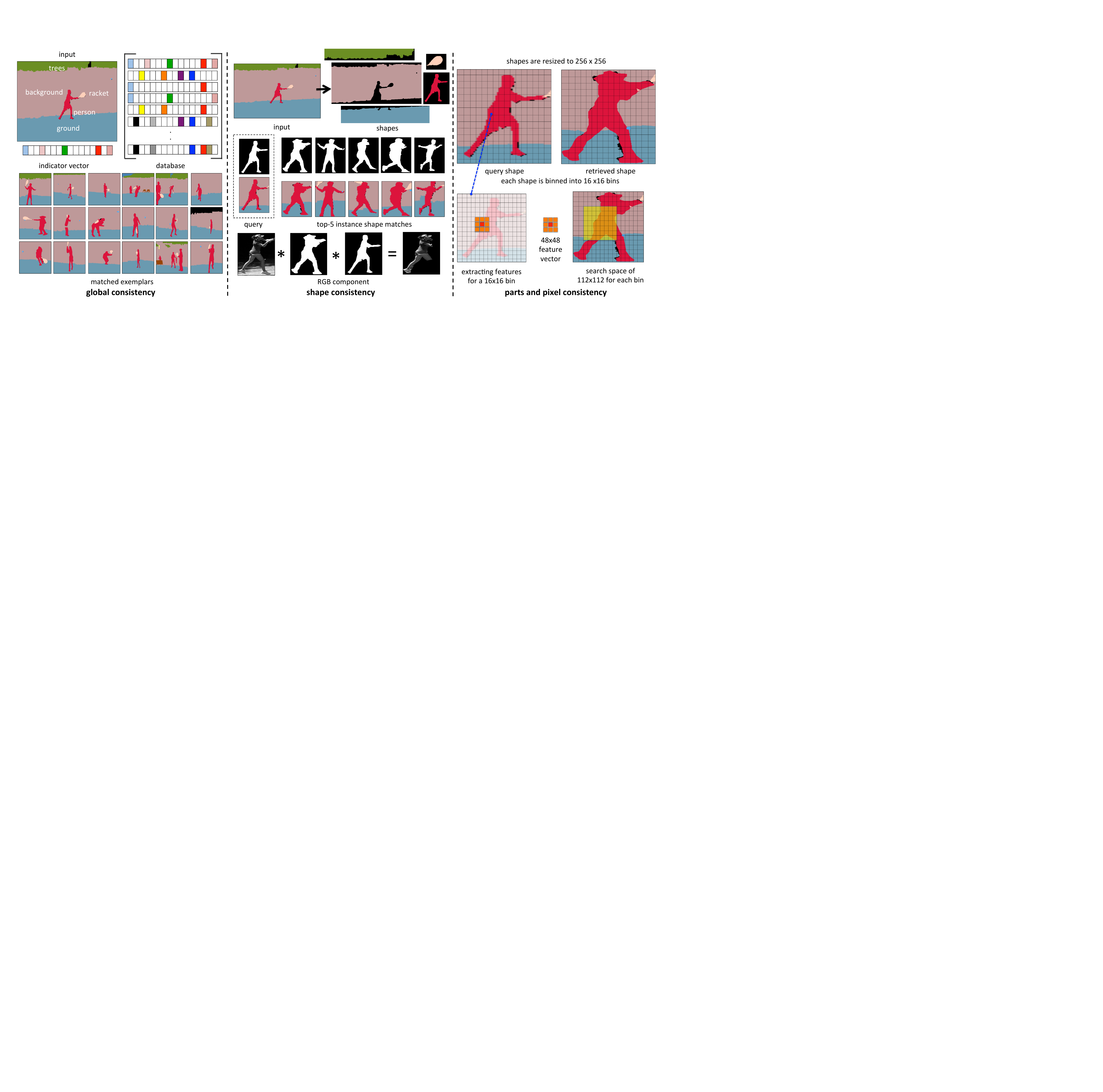}
  \caption{\textbf{Four stages of non-parametric matching: }  (1) \textbf{Global Consistency} - Given a semantic label map, we using an indicator vector that tells what categories are present. This indicator vector helps us to quickly find the relevant examples from database. This prunes away $99\%$ of exemplar database to get required shapes. (2) \textbf{Shape Consistency} - We extract various shapes from the input label mask, and match query shape to shapes within the exemplar set by using a shape-and-context feature. We show examples of top-$5$ retrieved shapes for a query shape on left. The image information from the retrieved shapes is then extracted  by considering the mask of query shape and retrieved shape; (3)  \textbf{Part Consistency} - We observe that global shape retrieved in last stage is missing information about the hands and legs of the query shape (human in this case). We define a local shape matching approach that looks in the neighborhood to synthesize parts. The query and top-$k$ shapes are resized to $256{\times}256$, and binned into $16{\times}16$ bins with each bin being a $16{\times}16$ patch. Each patch is represented by label information contained in it, and an additional $8$ neighboring patches. This provides contextual information about the surroundings. The parts are looked in an adjacent $112{\times}112$ region and the ones with minimum hamming distance is considered. \textbf{Pixel Consistency} algorithm follows a similar setup. See section~\ref{sec:approach} for more details.}
  \label{fig:overview}
\end{figure*}

Our work combines various ideas on shapes, deformable parts, context, and non-parametric approaches developed in last two decades. We position each separately and the specific insights for their particular usage.

\noindent\textbf{Shapes:}  Shapes~\cite{Koenderink:1990:SS:77527,marr1982vision,Roberts65}  emerge naturally in our world due to its compositional structure. If we had an infinite data-source with all potential shapes for all the objects, then our world could be represented by a linear combination of different shapes~\cite{gupta2010,Roberts65}. In this work, we aim to generate images from semantic and instance label maps as input. Meaningful shapes and contours~\cite{belongie2002shape,hariharan11} makes for an obvious interpretation for such an input. 

\noindent\textbf{Non-parametric approaches:} In an unconstrained in-the-wild data distribution consisting of both rigid and non-rigid objects, it becomes non-trivial to model such shapes for a one-to-many mappings. We, therefore, want to leverage the shape information explicitly from the training data in our formulation by simple \emph{copy-pasting}. Non-parametric methods~\cite{Efros03,Efros:2001,Efros99,Freeman2002,Hertzmann:2001,johnson11cg2real} find their use for various computer vision tasks such as texture-synthesis~\cite{Efros:2001,Efros99}, image super-resolution~\cite{Freeman2002}, action recognition~\cite{Efros03}, or scene completion~\cite{Hays:2007}. 

\begin{figure}
  \centering
  \includegraphics[width=\linewidth]{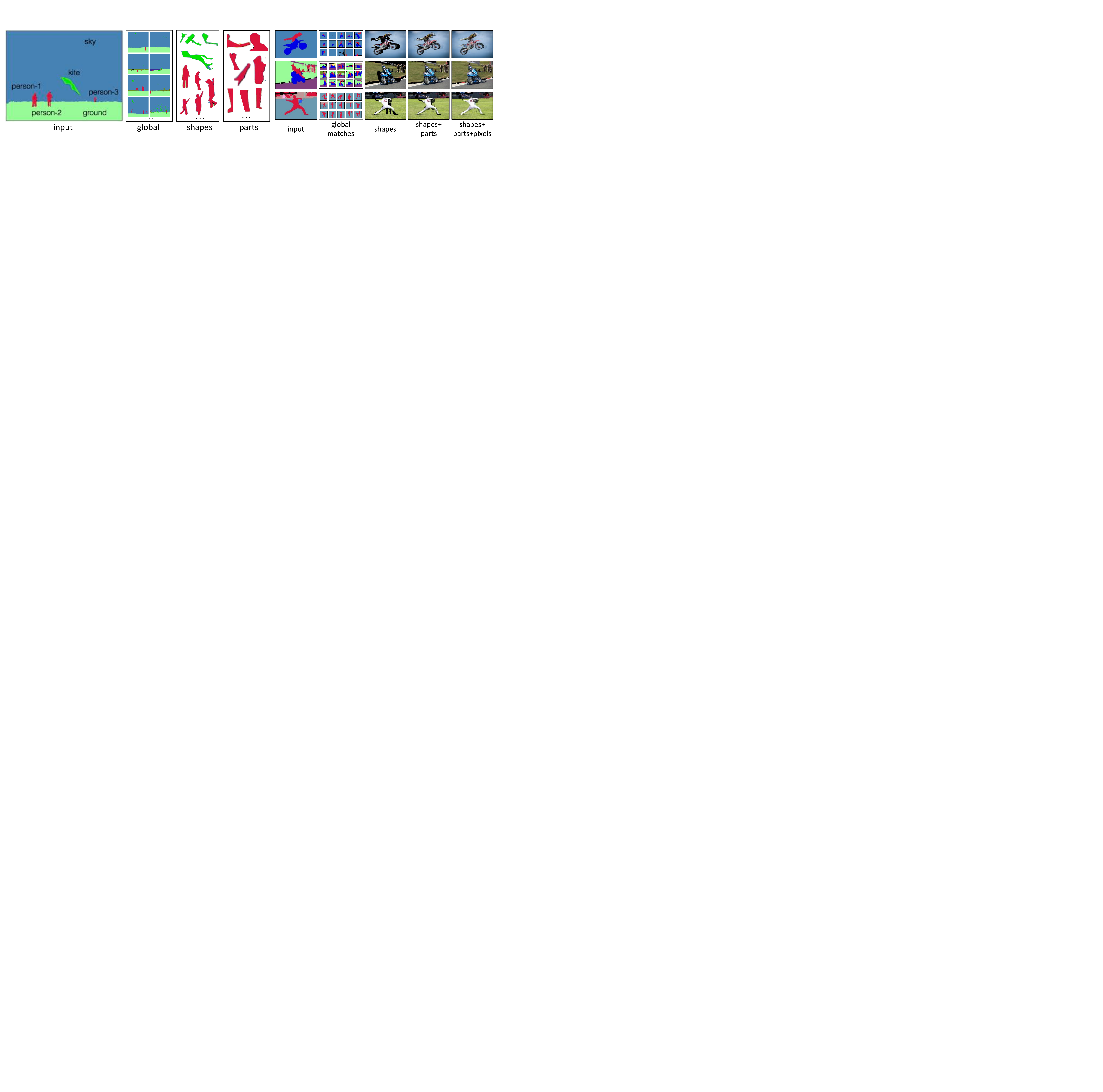}
  \caption{\textbf{Hierarchical Composition: } Given an input label map (left),  we  show the outputs of four stages of our non-parametric matching approach. The second column shows the top-$15$ global matches for the input. The third column shows the output of composition of global shapes extracted using our shape consistency algorithm. The fourth column shows the improved output by introducing local part consistency to previous output. Finally, minute pixel level holes are filled by our pixel-consistency algorithm.}
  \label{fig:composition}
\end{figure}

\noindent\textbf{Scene Composites:} Our work draws similarity to idea of scene composites~\cite{SceneCollaging,sims2018,russell2006using,russell2009segmenting}. Russell et al~\cite{russell2009segmenting} use shapes or scene composites to query matches for semantic segmentation using LabelMe dataset~\cite{russell2008}. In an another work, Russell et al~\cite{russell2006using} use similar idea of composites for object discovery. Isola and Liu~\cite{SceneCollaging} use this idea of composites for scene parsing and collaging. Recently, Qi et al~\cite{sims2018} used shapes in a semi-parametric form to synthesize images from a semantic label map. These different approaches~\cite{sims2018,russell2009segmenting} on scene composites or shapes are however constrained to rigid and non-deformable objects from a constrained data-distribution such as road-side scenarios from LabelMe~\cite{russell2008}, or Cityscapes~\cite{Cordts2016Cityscapes}. Our work extends the prior work to non-rigid and deformable shapes from an unconstrained in-the-wild data distribution. Figure~\ref{fig:concerns}-(c) shows how global shapes are insufficient and one needs to consider local information about parts and pixels.

\noindent\textbf{Deformable Objects \& Parts:} The global shape fitting can be reliably estimated for non-deformable objects but local shapes or \emph{parts}~\cite{Biederman87,FelzenszwalbMR_CVPR_2008,gu2009recognition} are required when considering non-rigid objects. The prior work on local components~\cite{Biederman87}, regions~\cite{gu2009recognition}, or parts~\cite{FelzenszwalbMR_CVPR_2008,Singh2012DiscPat,Yang:2013} has largely been focused on recognition. On the other hand, our work draws insight from ideas on compositional matching~\cite{BoimanI06,Faktor2013} and we use the parts, components, and regions for synthesizing images. In this work, we relax strict shape matching to do local part generation from various global shapes to do image synthesis. This enables us to consider local information without any explicit part labels.

\noindent\textbf{Context:} Context is a natural and powerful tool to put things in perspective~\cite{biedermansemantics,hoiem2006}. There is a wide literature on the use of context in computer vision community~\cite{Divvala-2009-10228,mottaghi_cvpr14} and is beyond the scope of this work to illustrate them completely. In this work, the contextual information comes for free with our input, i.e. semantic label map. We use this context at both global and local level to do better and faster matching of global shapes, parts, and pixels. The contextual information, while itself remaining in background, enables us to do an effective non-parametric matching.

\noindent\textbf{User Control:} Multiple works~\cite{pixelnn,Barnes:2009,lalonde-siggraph-07,wang2017high,zhang2017real,zhu2016generative} in computer graphics and vision literature have demonstrated the importance of user-controlled image operations. Grab-cut~\cite{Rother:2004} enables user-based segmentation of a given scene. Lalonde et al~\cite{lalonde-siggraph-07} use a non-parametric approach to insert objects in a given image. Kholgade et al~\cite{OM3D2014} demonstrate a user-controlled 3D object manipulation. In this work, we demonstrate how shapes can be used naturally and intuitively for a user controllable content creation and manipulation.


\section{Method}
\label{sec:approach}

Given a semantic and an instance label map, $X$, our goal is to synthesize a new image, $Y$. Our formulation is a hierarchical non-parametric matching ensuring the following stages in order: (1) global scene consistency; (2) instance shape consistency; (3) local part consistency; and finally (4) minute pixel-level consistency. Figure~\ref{fig:overview2} gives an overview of our approach.

\begin{figure*}[t]
  \centering
  \includegraphics[width=\linewidth]{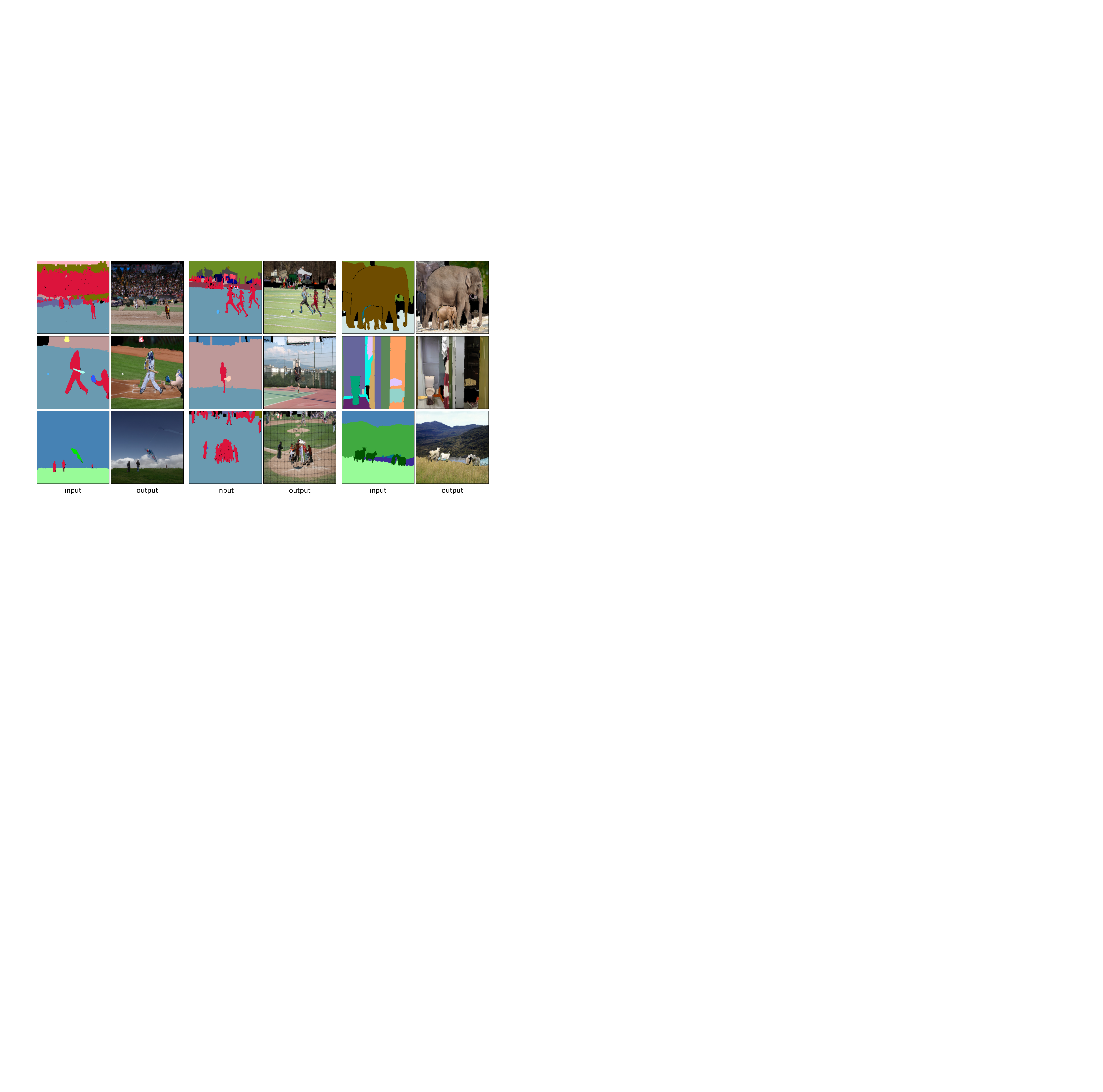}
  \caption{\textbf{Non-parametric matching:} Our approach for generating images from a semantic label map by non-parametric matching of global shapes, local parts, and pixel-consistency. The above examples contain varying background, cluttered environment, varying weather and illumination conditions, and multiple rigid and non-rigid objects in various shapes and forms.}
  \label{fig:non_parametric}
\end{figure*}

\noindent\textbf{Global Scene Consistency:} In a big data setting with hundred thousands and million examples, doing nearest neighbors could be a time-consuming process. We make this process faster by using a global indicator vector ( or scene context) to prune the list of training examples from which the shapes should be extracted. Only those examples are considered if they fall in one of three categories: (1) their global image has same labels as input; (2) the labels in input is its subset; (3) the labels in input is its superset. This reduces the search space from hundred thousand shapes to a few hundreds. We further prune them to top-$N$ images for searching shapes by  computing a global coverage and a pixel coverage score. 

A \textbf{global coverage} score is computed to ensure the top-$N$ label maps in the training set have similar distribution of labels as are in a given query label map. We compute the normalized histogram of labels (both query and training), and compute a $l_2$ distance between query and training label map. A \textbf{pixel coverage} score is computed to ensure we select the images with maximum pixel-to-pixel overlap. This score is computed by aligning a query label map and an example from training set, followed by the hamming distance between them. To make it faster, we resize the images to $100{\times}100$ and then compute the normalized hamming distance between the respective labels. We sum both global coverage and pixel coverage scores, and choose $N$ images in the training set with the lowest scores. This use of global scene context drastically reduces the search space for our non-parametric approach, and enables to do synthesis on everyday computing devices (single core CPUs instead of GPUs). Figure~\ref{fig:overview}-(left) shows similar examples retrieved by simple matching of indicator vector.

\noindent\textbf{Instance Shape Consistency:} We seek shapes as the first step to define different components in an image. We represent the shapes in an instance and semantic label mask as $\{x_1, x_2, ..., x_N\}$  where $N$ is the total number of shapes for a given input. Each shape has an associated semantic label $l:l \in \{1, 2, ..., L\}$ where $L$ is the number of unique labels.  We then make a tight bounding box around a shape $x_i$ so that it could be used as a rectangular convolutional filter ($w_i$) to retrieve similar shapes within the exemplar set. The filter is fixed to $50{\times}50$ in this work. We represent a  filter/shape using: (1) a simple \textbf{mask operator}: the part of a shape ($x_i$) in the filter ($w_i$) is set to $1$, and the remaining part is set to $-1$. This enables to match the shapes with similar boundaries and details; and (2) a \textbf{contextual operator}: we extract the labels from the input label mask for this filter. This information helps in getting better matches with similar context. 

We use the mask operator ($m$, where $m_{i}[k] \in \{-1,1\}$) and contextual operator ($v$, where $v_{i}[k] \in \{1,2,...,N_{c}\}$ and $N_{c}$ is number of categories) to match a query shape with shapes in exemplar set using the scoring function: 

\begin{align}
S_{shape}\Big((m_i, v_i), (m_j,v_j)\Big) =  \sum_{t}  m_{i}[t].m_{j}[t] + \nonumber \\
I(v_{i}[t] = v_{j}[t]), 
\label{eq:shape}
\end{align}

where $I$ is an indicator function with value $1$ when true, and zero otherwise. We ignore the shapes from exemplar set when the ratio of their aspect-ratio to that of query shape is either less than $0.5$ or greater than $2$. Using a fixed size filters and low-res label masks help us to generate composite of arbitrarily high resolution without any extra computational cost. The RGB component for an extracted shape is its intersection with the query shape, i.e. only pixels active in both extracted and query shape are considered. Figure~\ref{fig:overview}-(middle) illustrates this part of our algorithm.

\begin{figure*}[t]
  \centering
  \includegraphics[width=\linewidth]{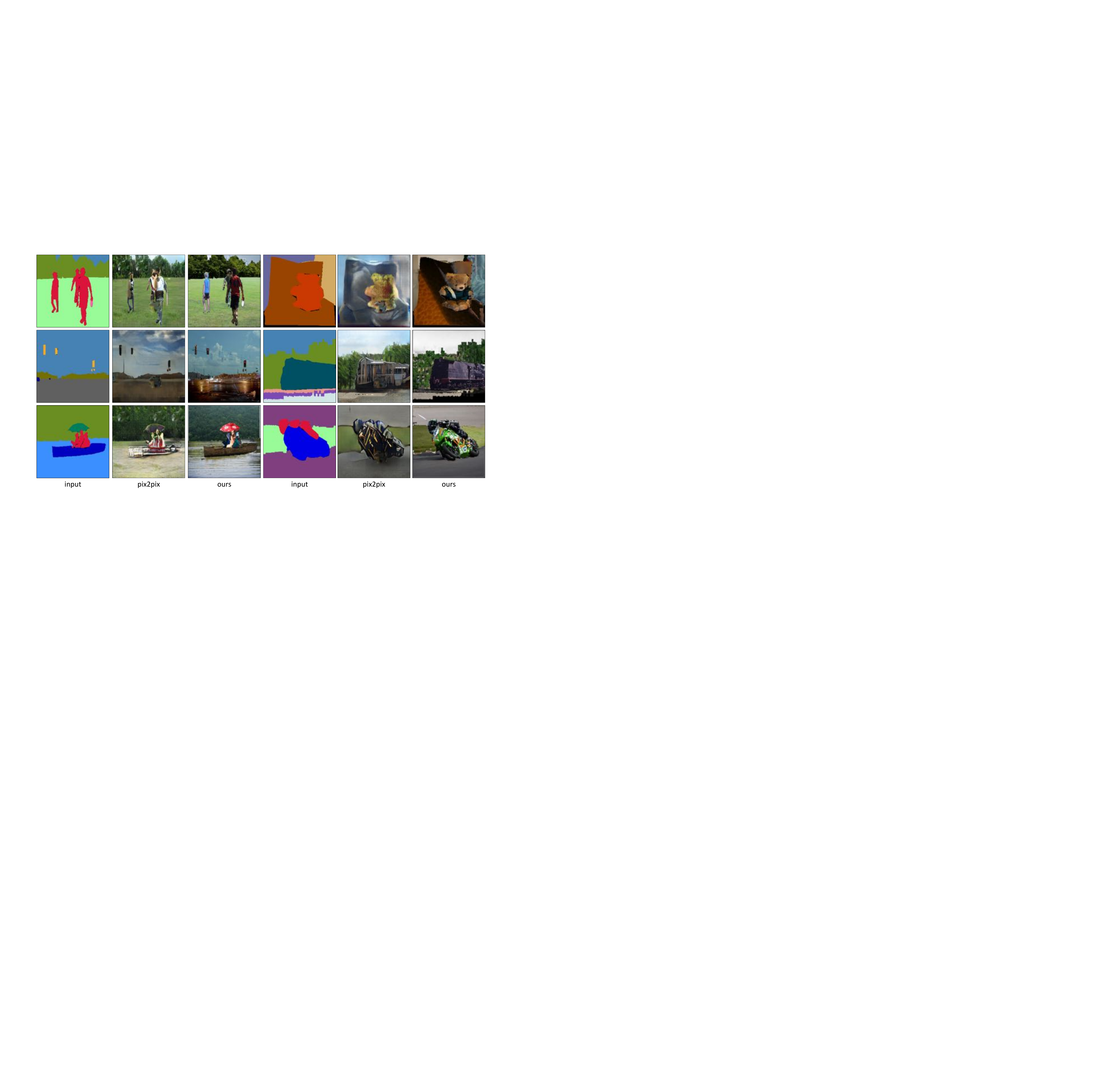}
  \caption{\textbf{Parametric vs. Non-parametric:} We generate images from an input label map. The second column shows the output of Pix2Pix~\cite{pix2pix2016} trained on COCO training set. The third  column shows the output of our non-parametric hierarchical matching approach.}
  \label{fig:qualitative}
\end{figure*}

\noindent\textbf{Local Part Consistency: } Modeling occlusions and deformable aspects of non-rigid objects in real world are extremely hard. The problem is even aggravated with noisy shape inputs. The insufficient shape data and non-rigid objects in real world leads to parts and local regions~\cite{Biederman87}. We seek parts from top-$k$ global shapes. Importantly, the part information is required when a global shape is not able to capture. We extract the knowledge of parts from the global shapes in a spirit similar to non-parametric texture synthesis~\cite{Efros99}. The shape components are resized to $256{\times}256$, so that local information can be well searched. We extract a $16{\times}16$ patch from the resized global shape template. Local contextual information (similar to HOG~\cite{dalal2005histograms}, or Group-Normalization~\cite{WuH18}) is used by considering the neighboring $8$ patches. A patch ($p$) is, therefore, represented by a $256{\times}9$ dimensional vector containing the label information in its region and surroundings. The parts are scored using: 

\begin{align}
S_{part}(p_i, p_j) =   \sum_{t} I(p_{i}[t] = p_{j}[t])
\label{eq:parts}
\end{align}

Importantly, we do not need to look in a larger window for part matching as we have weakly aligned global shapes. Therefore, we restrict the patches to look in a surrounding $5{\times}5$ patch window. This corresponds to $112{\times}112$ pixel window in a resized global shape template. We copy the RGB component from the best matching patch window. Figure~\ref{fig:overview}-(right) shows the part of our algorithm to compute part-consistency.

\noindent\textbf{Minute Pixel Consistency: } The shapes and parts have accounted for most of the non-parametric image synthesis. However, they does not ensure pixel-level consistency and often ends up with minor holes in an image. We enforce a pixel-level consistency in this process to account for the remaining holes in synthesized image. This process is similar to our part consistency algorithm, except that it is done on every pixel. Each pixel is represented by a surrounding $11{\times}11$ window. We use the criterion in Eq.~\ref{eq:parts} to compute similarity between two feature vectors. To expedite this matching, we compute features for a low-res input label map ($128{\times}128$) as pixel consistency is ensured to fill minor holes alone.  Finally, we look in surrounding region of $5{\times}5$ from a $128{\times}128$ image to fill in the information as global and local consistency have already been accounted by shape and part consistency.

\noindent\textbf{Hierarchical Composition: } We combine the information hierarchically from shapes, parts, and pixels to generate a full image. Figure~\ref{fig:composition} shows the composition starting from an input label map. Firstly, we find relevant examples using global scene context. We then use the instance shape components to fill the major chunk of image. The missing information is then filled using the local part consistency. Finally, the minor holes are filled using the pixel-level consistency. The combination of these four stages enable us to efficiently generate a image from an input label mask by simple non-parametric matching. 

\noindent\textbf{Qualitative Results: } Figure~\ref{fig:non_parametric} shows outputs generated by our approach for varying background, cluttered environment, varying weather and illumination conditions, and multiple rigid and non-rigid objects in various shapes and forms. Figure~\ref{fig:qualitative} contrasts our approach with Pix2Pix~\cite{pix2pix2016}.

\begin{figure*}[t]
  \centering
  \includegraphics[width=\linewidth]{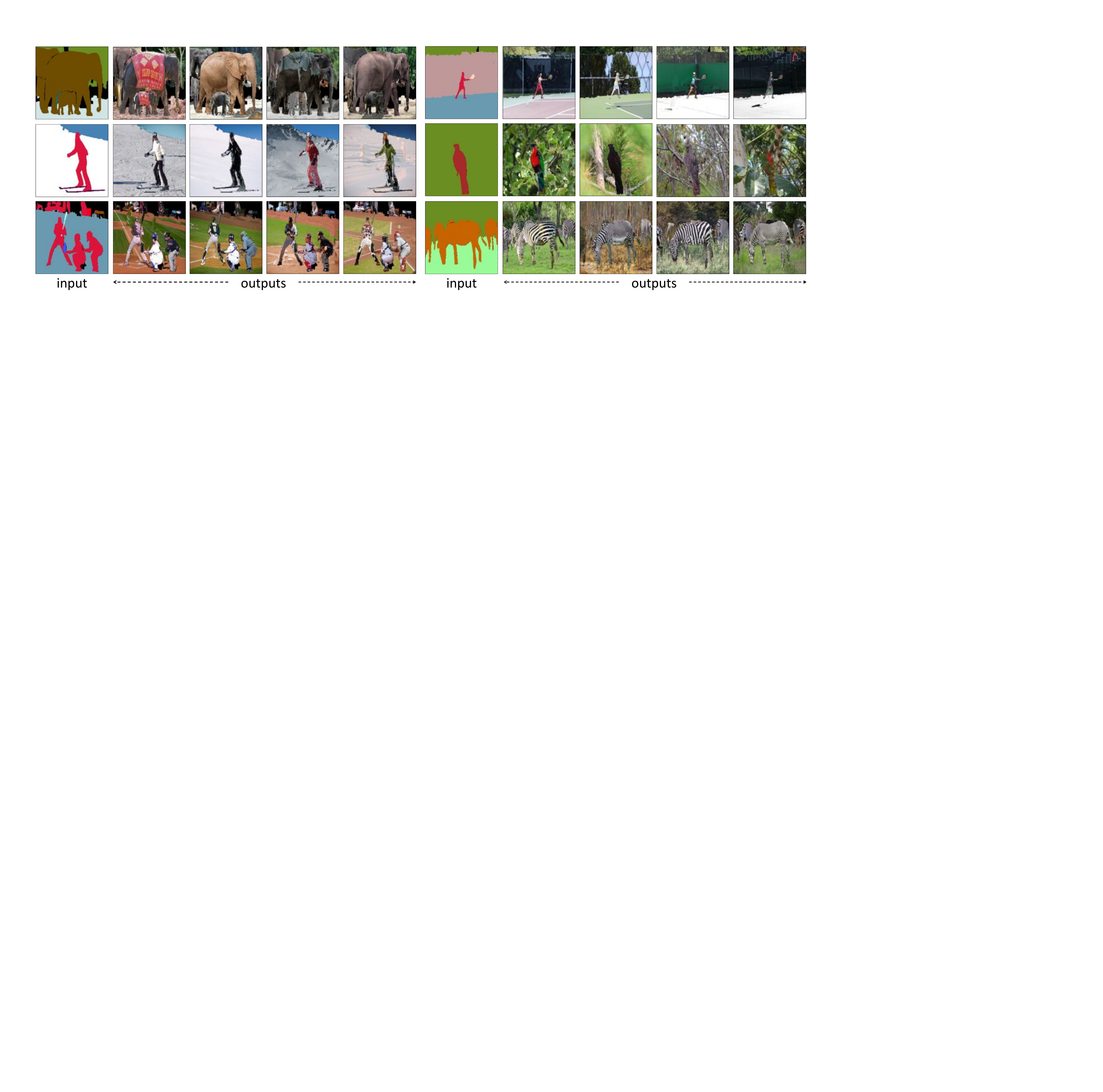}
  \caption{\textbf{Multiple Outputs: } Our approach can easily generate exponentially large number of outputs by changing shapes and parts. We show four outputs generated for each label map.}
  \label{fig:multi}
\end{figure*}

\noindent\textbf{Multiple-Outputs: } A salient aspect of considering shapes and parts in a non-parametric matching provides multiple outputs for free. We can combine various  extracted shapes in exponential ways without any extra overhead. We show multiple examples synthesized for a given label map using our approach in Figure~\ref{fig:multi}. Generating these multiple outputs is not trivial when using parametric approaches~\cite{chen2017photographic,pix2pix2016}, and has been a subject of multiple studies~\cite{Ghosh2017MultiAgentDG,zhu2017toward}.

\begin{figure*}[t]
  \centering
  \includegraphics[width=\linewidth]{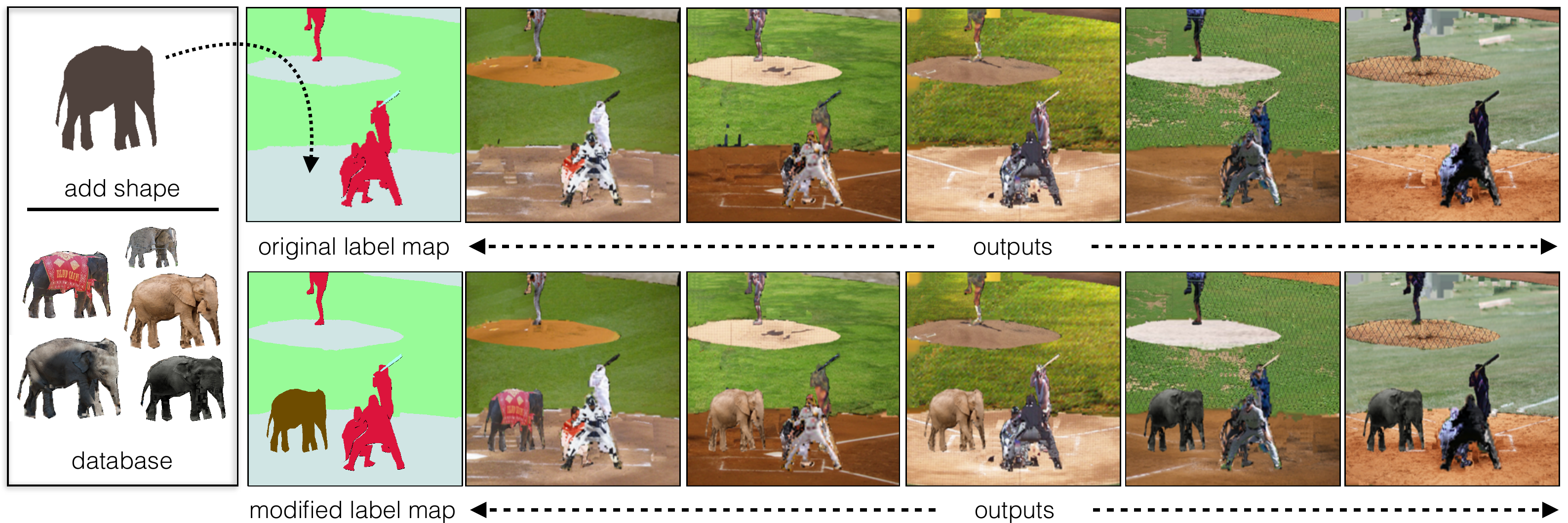}
  \caption{\textbf{User-Intervention \& Manipulation: }  The first row shows multiple outputs generated for a given label map. We insert the shape of \emph{elephant} in this input label map to synthesize an \emph{elephant on the baseball field}. The second row shows multiple outputs for the modified label map.}
  \label{fig:img_manipulation}
\end{figure*}

\noindent\textbf{User Control: } We finally demonstrate the applicability of our approach for a user controllable content creation in Figure~\ref{fig:img_manipulation}. Even though it is rare to see \emph{an elephant on the baseball field}, our approach can easily generate such an example by inserting shapes. More importantly, synthesis and manipulation aspects go hand-in-hand for our approach. A human user can clearly \emph{interpret} and \emph{influence} any stage of synthesis, and can easily generate a different output by varying a shape. Manipulation naturally emerges in our non-parametric approach without any additional efforts.


\section{Experiments}
\label{sec:experiments}

\noindent\textbf{Dataset: } We use semantic and instance label mask from COCO~\cite{MSCOCO-2014} to study the problem of in-the-wild image synthesis and manipulation. This dataset consists of $134$ different objects and stuff categories making it one of the most diverse and varied publicly available dataset.  There are $118,287$ images in the training set ($40\times$ more than Cityscapes~\cite{Cordts2016Cityscapes}), and $5,000$ images in the validation set ($100\times$ more than Cityscapes). We use the paired data of labels and images from training set to extract global shapes and synthesize parts and pixels. The images are synthesized using semantic and instance label masks in validation set. Our approach does not require any training, and therefore can use the labels and image component from anywhere. For the sake of fair comparison with parametric approaches, we restrict ourselves to COCO training data. 

\noindent\textbf{Baselines: } To the best of our knowledge, there does not exist a non-parametric approach that has attempted the problem of the in-the-wild image synthesis from label maps on this large scale. We, therefore, compare our approach with parametric approaches: (1). Pix2Pix~\cite{pix2pix2016}; and (2). Pix2Pix-HD~\cite{wang2017high}, using their publicly available codes. The complexity, diversity, and size of this dataset  makes it a computational challenge for a generative parametric approach to deal with. Training a vanilla Pix2Pix model took 20 days on a single Nvidia Titan-X GPU. On the same compute, we trained a Pix2Pix-HD model  for a month but did not observe any convergence. It may be possible that a reasonable Pix2Pix-HD model be trained if we let the training go longer for an extra month or two, or use advanced computational resources. It may also be due to design of architecture and hyper-parameters specifically suited for Cityscapes, and that efforts are required to tune hyper-parameters to make it work for a large and diverse dataset as COCO.  It is for this reason we also use Cityscapes to contrast our approach with prior works~\cite{chen2017photographic,pix2pix2016,sims2018,wang2017high} for the sake of fair comparison. Additionally, we resize our generated outputs to  $256 \times 256$ just to make a fair comparison with Pix2Pix on COCO. However, we can generate outputs having same resolution as that of input label masks without any increase in compute.

\noindent\textbf{FID Scores}: We compute FID scores~\cite{NIPS2017_7240} using the images generated from different approaches. Lower FID values suggest more realism. Table~\ref{tab:shapes} contrast FID scores computed on generated images (COCO) with Pix2Pix and Pix2Pix-HD (resized to $256{\times}256$ and $64{\times}64$ resolution). Without using any oracle, the top-$1$ example generated from our approach significantly outperforms the prior work. Additionally, note the performance improvement due to each stage in our hierarchical composition.

\begin{table}[h]
\scriptsize{
\setlength{\tabcolsep}{3pt}
\def\arraystretch{1.3}
\center
\begin{tabular}{@{}l c c c c c}
\toprule
\textbf{Method}    & \#examples	& Oracle &  FID score  			& FID score    \\
			   & 			 & 		& ($256{\times}256$)	& ($64{\times}64$) \\		
\midrule
Pix2Pix~\cite{pix2pix2016}		 &		1	& \xmark	& 	70.43 & 41.45 \\
Pix2Pix-HD~\cite{wang2017high}  	 &		1	& \xmark	&     157.13 & 109.49\\
\midrule
Ours (shapes)  &		1	& \xmark & 37.26 &  23.22\\
Ours (shapes+parts)  &		1	& \xmark & 32.62 & 18.02 \\
Ours (shapes+parts+pixels)		 &		1	& \xmark	& \textbf{31.63} & \textbf{16.61}\\
\bottomrule
\end{tabular}
\caption{\textbf{FID Scores on COCO:} We  compute FID score~\cite{NIPS2017_7240} to contrast the realism in outputs produced by different approaches. \textbf{Lower FID values suggest more realism}. We observe that our approach outperforms prior approaches significantly. We also demonstrate as how different stages in our hierarchical composition leads to better outputs.}
\vspace{-0.3cm}
\label{tab:shapes}}
\end{table}

\noindent\textbf{Mask-RCNN Scores: } We use a pre-trained Mask-RCNN~\cite{he2017maskrcnn} to study the quality of synthesis on COCO~\cite{MSCOCO-2014} for Pix2Pix and our approach. This model is trained for 80 object categories of COCO dataset.  While it is trained for instance segmentation, we use its output and convert it to semantic labels for consistency in evaluation. Our goal is to observe if we can get the same class labels from the synthesized images as one would expect from a real image. We, therefore, run it on original images from the validation set and use these pseudo semantic labels as ground truth for evaluation. Next, we run it on synthesized images and contrast it with the labels from original image. To measure the performance, we use three criterion: (1) mean pixel accuracy (PC); (2) mean class accuracy (AC); (3) mean intersection over union (IoU). Higher the score for each of the criterion, better is the quality of synthesis. Table~\ref{tab:mask-rcnn} contrasts the performance of our approach with Pix2Pix and demonstrates substantially better results. Our performance improves when an oracle is used to select the best from five outputs. Note that top-$100$ exemplar matches are used for instance shape matching in this experiment. 

\begin{table}[h]
\scriptsize{
\setlength{\tabcolsep}{3pt}
\def\arraystretch{1.3}
\center
\begin{tabular}{@{}l c c c c c c}
\toprule
\textbf{Method}    & \#examples	& Oracle &  PC & AC & IoU \\
\midrule
\textbf{Parametric}  &			& 	& 	&  &  \\
Pix2Pix~\cite{pix2pix2016}		 &		1	& \xmark	& 17.9	& 8.9 & 4.9 \\
\midrule
\textbf{Non-Parametric}  &			& 	& 	&  &  \\
Ours  &		1	& \xmark	& 44.5	& 31.0  & 20.9\\
Ours   &		5	& \checkmark	& 58.2 	& 41.2  & 31.4 \\
\bottomrule
\end{tabular}
\caption{\textbf{Mask-RCNN Scores on COCO: } We use a pre-trained Mask-RCNN model~\cite{he2017maskrcnn} to study the quality of image synthesis. We run it on synthesized images and contrast it with the labels from original image. To measure the performance, we use three criterion: (1) mean pixel accuracy (PC); (2) mean class accuracy (AC); (3) mean intersection over union (IoU). \textbf{Higher the score for each of the criterion, better is the quality of synthesis.} We outperform Pix2Pix. The performance further improves significantly when an oracle is used to select from five examples. }
\vspace{-0.3cm}
\label{tab:mask-rcnn}}
\end{table}

\noindent\textbf{Human Studies: } We did human studies on a randomly selected 500 images. We show the outputs of Pix2Pix, Pix2Pix-HD, and our approach (randomly picked one output from multiple) to human subjects for as much time as they need to make a decision. We asked them to pick one that looks close to a real image. The users were advised to use `none of these' if all approaches are consistently bad. $51.2\%$ times user picked an output generated from our approach, $7.8\%$ times the outputs from Pix2Pix, and preferred `none of these' $41\%$ times. The human studies suggest that while our approach is most likable, there are still many situations where our approach produced undesirable outputs.

\noindent\textbf{Cityscapes: } Table~\ref{tab:cityscapes} contrasts the performance of our approach with prior approaches~\cite{chen2017photographic,pix2pix2016,sims2018,wang2017high} that have specifically demonstrated on Cityscapes. Except Pix2Pix, we used publicly available results for this evaluation. Our approach is competitive to prior parametric and semi-parametric approaches with just $25$ exemplar matches to extract shapes and parts to compose a new image from semantic labels. The performance improves when using an oracle to select best amongst the $5$ generated outputs. Our performance may further improve as we increase the number of global images to do shape and part extraction.

\begin{table}
\scriptsize{
\setlength{\tabcolsep}{3pt}
\def\arraystretch{1.3}
\center
\begin{tabular}{@{}l c c c c c c}
\toprule
\textbf{Method}    & \#examples	& Oracle &  PC & AC & IoU \\
\midrule
\textbf{Parametric}  &			& 	& 	&  &  \\
Pix2Pix~\cite{pix2pix2016}		 &		1	& \xmark	& 72.5	& 29.5 & 24.6 \\
CRN~\cite{chen2017photographic}  &		1	& \xmark	& 49.0	& 22.5  & 18.2 \\
Pix2Pix-HD~\cite{wang2017high}  &		1	& \xmark	& 79.0	& 43.3  & 37.8 \\
\midrule
\textbf{Semi-Parametric}  &			& 	& 	&  &  \\
SIMS~\cite{sims2018}  &		1	& \xmark	&  68.6	& 35.1  & 28.1	\\
\midrule
\textbf{Non-Parametric}  &			& 	& 	&  &  \\
Ours (top-$25$)  &		1	& \xmark	& 67.1	& 38.0  & 30.5\\
Ours  (top-$25$) &		5	& \checkmark	& 71.3 	& 39.6  & 32.4 \\
\bottomrule
\end{tabular}
\caption{\textbf{PSP-Net Scores on Cityscapes: } We use a pre-trained PSP-Net model~\cite{zhao2017pspnet} to evaluate the quality of synthesized images ($1024{\times}2048$). This model is trained for semantic segmentation on cityscapes. We run the synthesized images through this model, and generate a semantic label map for each image. The semantic label map from the synthesized images is contrasted with the semantic label map from the original image. We compute three statistics for each approach: (1) Mean Pixel Accuracy (PC); (2) Mean Class Accuracy (AC); (3) Mean intersection over union (IoU). For each of these criterion- \textbf{higher the score, better is the quality of synthesis}. With just $25$ exemplar matches to extract shapes and parts, our non-parametric approach is competitive to the parametric models and semi-parametric models.}
\vspace{-0.3cm}
\label{tab:cityscapes}}
\end{table}

\section{Discussion \& Future Work}
\label{sec:discussion}

We present an exceedingly simple non-parametric approach for image synthesis and manipulation in-the-wild. While the diverse data-distribution and large datasets make it challenging for parametric approaches to operate on, it enables simple matching of shapes and parts to work well. The non-parametric matching enables us to generate exponentially large number of outputs by varying shapes and parts. Importantly, shapes and parts are intuitive to a normal human user as well. This makes our approach interpretable and suitable for user-controllable content creation and editing. The future work in this direction may address smarter ways of combining shapes and parts information, and explore spatiotemporal consistency to do in-the-wild video synthesis and manipulation.

{\small
\bibliographystyle{ieee}
\bibliography{references}
}

\end{document}